\documentclass[10pt,twocolumn,letterpaper]{article}

\usepackage{cvpr}              

\usepackage[dvipsnames]{xcolor}

\usepackage{graphicx}
\usepackage{tablefootnote}
\usepackage{amsmath}
\usepackage{amssymb}
\usepackage{booktabs}
\usepackage{colortbl}
\usepackage[dvipsnames]{xcolor}
\definecolor{tablegray}{rgb}{0.8, 0.8, 0.8}

\usepackage{listings}
\usepackage{multirow}
\usepackage{algorithm}
\usepackage{algpseudocode}
\usepackage{tikz}
\newcommand{\xmark}{%
\tikz[scale=0.23] {
    \draw[line width=0.7,line cap=round] (0,0) to [bend left=6] (1,1);
    \draw[line width=0.7,line cap=round] (0.2,0.95) to [bend right=3] (0.8,0.05);
}}
\usepackage[normalem]{ulem}

\newcommand{\customparagraph}[1]{\smallskip\noindent\textbf{#1}\quad}

\definecolor{backcolour}{rgb}{0.95,0.95,0.92}

\lstdefinestyle{mystyle}{
    backgroundcolor=\color{backcolour},   
    basicstyle=\ttfamily,
}

\newcommand\blfootnote[1]{%
  \begingroup
  \renewcommand\thefootnote{}\footnote{#1}%
  \addtocounter{footnote}{-1}%
  \endgroup
}

\definecolor{cvprblue}{rgb}{0.21,0.49,0.74}
\usepackage[pagebackref,breaklinks,colorlinks,citecolor=cvprblue]{hyperref}

\def\paperID{20} 
\def\confName{CVPR}
\def\confYear{2025}

\title{Moment Sampling in Video LLMs for Long-Form Video QA}

\author{%
  \textbf{Mustafa Chasmai$^{1}$\thanks{Work done during internship at Dolby Laboratories.} \quad Gauri Jagatap$^2$ \quad Gouthaman KV$^2$} \\ \textbf{Grant Van Horn$^1$ \quad  Subhransu Maji$^1$ \quad Andrea Fanelli$^2$} \\
  $^1$University of Massachusetts Amherst \quad $^2$Dolby Laboratories \\
  {\small \texttt{\{mchasmai,gvanhorn,smaji\}@umass.edu}\ \  \texttt{\{Gauri.Jagatap,Gouthaman.KV,Andrea.Fanelli\}@dolby.com}}
}

\begin{document}
\maketitle

\begin{abstract}
    Recent
\blfootnote{Presented at the Workshop on Video LLMs, CVPR 2025. \href{https://www.crcv.ucf.edu/cvpr2025-vidllms-workshop/}{[link]}} advancements in video large language models (Video LLMs) have significantly advanced the field of video question answering (VideoQA). While existing methods perform well on short videos, they often struggle with long-range reasoning in longer videos. To scale Video LLMs for longer video content, frame sub-sampling (selecting frames at regular intervals) is commonly used. However, this approach is suboptimal, often leading to the loss of crucial frames or the inclusion of redundant information from multiple similar frames. Missing key frames impairs the model's ability to answer questions accurately, while redundant frames lead the model to focus on irrelevant video segments and increase computational resource consumption. In this paper, we investigate the use of a general-purpose text-to-video moment retrieval model to guide the frame sampling process. We propose ``moment sampling", a novel, model-agnostic approach that enables the model to select the most relevant frames according to the context of the question. Specifically, we employ a lightweight moment retrieval model to prioritize frame selection. By focusing on the frames most pertinent to the given question, our method enhances long-form VideoQA performance in Video LLMs. Through extensive experiments on four long-form VideoQA datasets, using four state-of-the-art Video LLMs, we demonstrate the effectiveness of the proposed approach.

\end{abstract}

\section{Introduction}
\label{sec:intro}

\begin{figure}[t]
    \centering
    \includegraphics[width=0.96\linewidth]{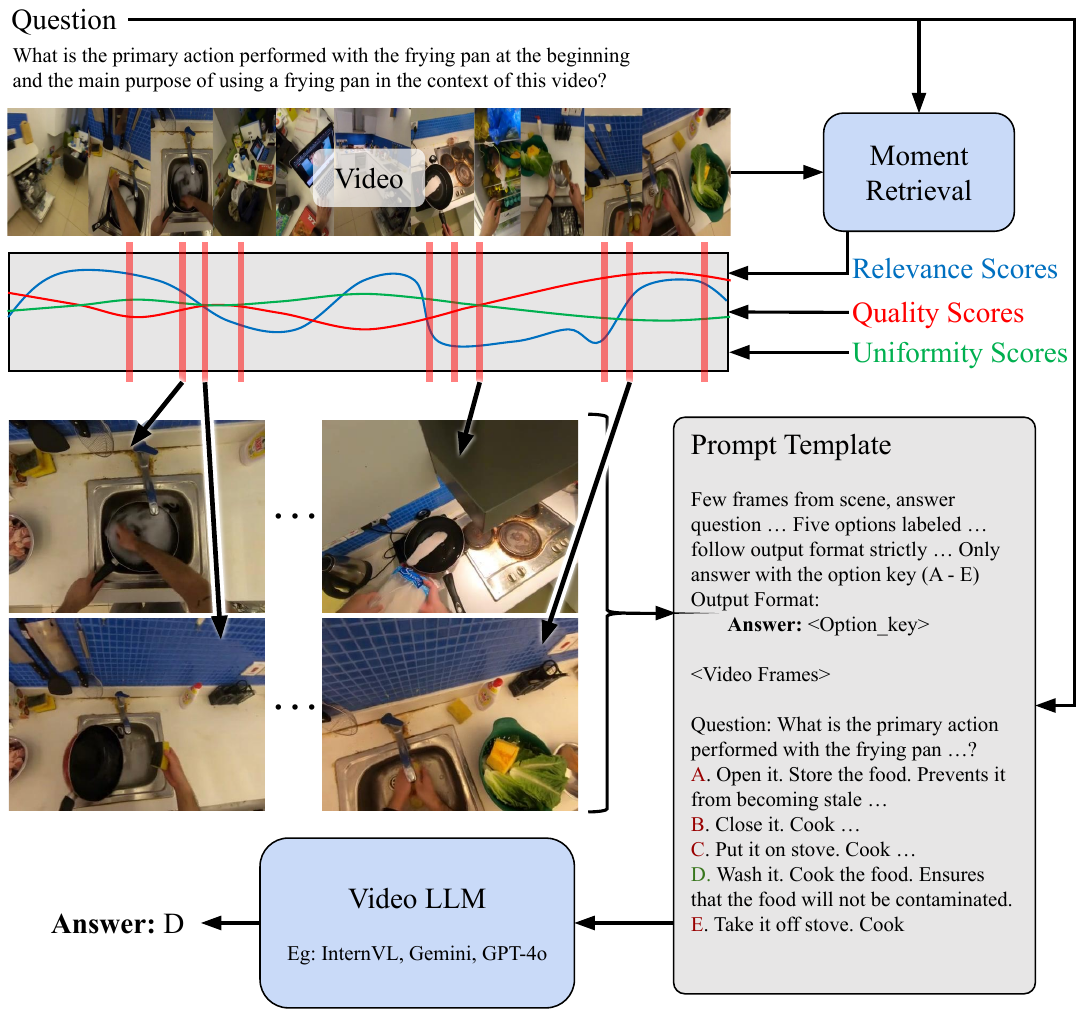}
    \caption{\textbf{Moment Sampling for VideoQA.} Given a video and a question (top), we first retrieve moments in the video that are relevant to answer the question. Retrieved moments, along with quality and uniformity scores are used to sample a few frames. These are given as input to a Video LLM to obtain the answer. }
    \label{fig:teaser}
    \vspace{-2mm}
\end{figure}

Video question answering (VideoQA) is a challenging and impactful task with numerous real-world applications, including egocentric assistants for wearables, conversational agents that interpret video content, intelligent tools for long-form video editing, educational aids in lecture videos, and efficient browsing of surveillance footage, etc. These scenarios often involve complex, untrimmed videos that require multimodal reasoning across visual, textual, and sometimes auditory information, within a shared context. To answer questions accurately, a model must identify salient objects and events, understand relationships between them, reason over time, and in many cases, apply common-sense or domain-specific knowledge, such as interpreting character interactions in a film or understanding mechanical concepts in an instructional video, etc.  

Recent advancements in multimodal foundation models, particularly Video Large Language Models (VideoLLMs), have significantly elevated the capabilities of VideoQA systems. These models are designed to integrate visual~\cite{liu2024visual, bai2023qwen, awadalla2023openflamingo, li2023blip}, auditory~\cite{yang2023uniaudio, kong2024audio}, and textual~\cite{touvron2023llama, dubey2024llama, achiam2023gpt, team2024gemma, jiang2023mistral} modalities into a unified reasoning framework. While current VideoLLMs have demonstrated strong performance on short video clips, typically ranging from a few seconds to a few minutes~\cite{jang2017tgif, lin2023video, cheng2024videollama, chen2024internvl, wang2024internvideo2}, scaling them to long-form VideoQA introduces new challenges. These include maintaining temporal coherence across extended durations and managing computational constraints imposed by the length and complexity of input videos.  To reduce the input size, existing approaches often resort to uniform sampling, where a fixed number of frames or segments are evenly sampled from the video (e.g., 100 frames from a 10-minute clip)~\cite{rawal2024cinepile, chen2024internvl, lin2023video}. However, this strategy is often suboptimal: it risks overlooking semantically important content while including redundant or irrelevant frames, ultimately limiting the model’s ability to perform accurate reasoning and increasing computational overhead.  Addressing these limitations is especially crucial for VideoLLMs, which are often sensitive to input quality due to token and context length constraints. As such, there is a pressing need for intelligent, question-aware frame sampling strategies that dynamically select frames based on their relevance to the given question. By aligning the input with the question’s semantic and temporal requirements, such methods can enhance both the efficiency and effectiveness of VideoLLMs in long-form VideoQA, paving the way for more scalable and capable systems in real-world video understanding.

Moment retrieval has emerged as another task in video understanding, focused on identifying and localizing the most relevant temporal segment within an untrimmed video that aligns with a given natural language query~\cite{lei2021detecting, lee2023bam, moon2023query, moon2023correlation, liu2022umt}. This task demands fine-grained temporal grounding, often across lengthy and complex video content. Recent advances in vision-language modeling have significantly enhanced the ability to align textual queries with specific video segments, enabling more accurate and context-aware moment retrieval. Given these strengths, moment retrieval models offer a compelling foundation for guiding question-aware frame sampling in long-form VideoQA.  

In this paper, we systematically explore how recent advances in moment retrieval can be leveraged to improve long-form VideoQA in VideoLLMs. To this end, we propose a model-agnostic approach called \textit{moment sampling}, which intelligently guides frame selection based on moment retrieval cues. Our method is compatible with any VideoLLM and does not require additional retraining or fine-tuning. An overview of the proposed framework is illustrated in Fig.~\ref{fig:teaser}.
We evaluate the effectiveness of \textit{moment sampling} across a diverse set of VideoLLMs, including proprietary models such as GPT-4o~\cite{openai_gpt4o} and Gemini 1.5 Pro~\cite{reid2024gemini}, as well as the open-source models such as VideoLLaVA~\cite{lin2023video}  and InternVL2~\cite{chen2024internvl}. Experiments are conducted on four public datasets: EgoSchema~\cite{mangalam2024egoschema}, CinePile~\cite{rawal2024cinepile}, NextQA~\cite{xiao2021next}, and IntentQA~\cite{li2023intentqa}. 

To the best of our knowledge, this is the first work to leverage advancements in moment retrieval for improving long-form VideoQA within the context of VideoLLMs. Our experiments demonstrate that the proposed Moment Sampling strategy significantly enhances frame sampling efficiency and consistently improves performance across all evaluated VideoLLMs and benchmark datasets.

\section{Related Work}
\label{sec:related}

\customparagraph{Long-form Video Understanding:} Video understanding showed significant advancements in the literature encompasses a wide range of tasks, such as action recognition~\cite{zhao2023learning, yan2022multiview, lee2024cast, wu2022memvit, kondratyuk2021movinets, herzig2022object, arnab2021vivit}, video captioning and question-answering~\cite{xu2023mplug, chen2023vast, chencosa, kuo2023mammut, ging2020coot}, summarization~\cite{gygli2014creating, apostolidis2021combining, zhu2022relational}, and retrieval~\cite{lei2021detecting, moon2023query, xue2023clip, guan2023pidro}. These methods often works well in the case of short-clips spanning a few seconds to minutes. Understanding long-form videos presents unique challenges that remain relatively underexplored. Unlike short clips, long-form videos require sophisticated methods for extended temporal reasoning, modeling event sparsity, and efficiently retrieving relevant information across lengthy sequences. Existing methods for long-form video understanding can be broadly categorized into two approaches: hierarchical strategies like Video ReCap~\cite{islam2024video}, which focus on summarizing and captioning long videos, and sequence-based models such as Mamba~\cite{gu2023mamba} and state-space architectures~\cite{gu2022efficiently}, designed to process and integrate extended temporal contexts~\cite{park2024videomamba, li2024videomamba}. 

\customparagraph{Long-form Video Question Answering:} VideoQA has emerged as a crucial benchmark for evaluating fine-grained video understanding, particularly in long-form scenarios. Traditional supervised methods train multimodal models on video-question-answer triplets~\cite{ko2023large, balazevic2024memory, papalampidi2024simple}. For example, FlippedVQA~\cite{ko2023large} not only trains models to predict answers but also to generate corresponding questions and reconstruct video content, requiring minimal fine-tuning of adapter layers. MC-ViT~\cite{balazevic2024memory} introduces a transformer architecture that scales effectively for long-context video understanding, while LongViViT~\cite{papalampidi2024simple} employs enhanced contrastive objectives for optimizing long-form VideoQA training. However, these approaches often rely on domain-specific models, and domain shifts, such as those between movie and egocentric videos, pose significant challenges for generalization across datasets.

\customparagraph{VideoLLMs for Long-form VideoQA:} To address the limitations of supervised and domain-specific approaches, zero-shot and prompt-based methods utilizing VideoLLMs have gained significant attention. These models can be broadly classified into two categories: those that first generate textual narrations through captioning or summarization models, and those that directly process frame-level video input. The former generates dense captions or summaries, which are subsequently queried by a pre-trained LLM to answer questions. For instance, VideoAgent~\cite{wang2024videoagent} uses an LLM as a central reasoning module, iteratively gathering relevant content to answer questions, while VideoTree~\cite{wang2024videotree} improves efficiency by scoring frame relevance and constructing an adaptive tree-based textual representation. While these methods offer scalable inference and generalization, they rely heavily on the quality of captioning models, which often miss crucial visual cues necessary for precise answers.

In contrast, VideoLLMs that operate directly on frame-level video input offer a promising alternative. These models typically down-sample long videos by selecting frames at regular intervals or segmenting them into shorter clips of just a few seconds. Proprietary models like GPT-4o~\cite{openai_gpt4o} and Gemini 1.5 Pro~\cite{reid2024gemini}, and open-source models such as Video-LLaVA~\cite{lin2023video}, InternVL2~\cite{chen2024internvl}, and Video-LLaMA~\cite{cheng2024videollama}, fall into this category. However, uniform frame sampling is inherently limited, as it may miss semantically important moments while including redundant or irrelevant content, which compromises the model’s accuracy and increases processing overhead. These challenges highlight the need for more intelligent and adaptive frame selection strategies. Recent efforts, such as LVNet~\cite{park2024too}, have made progress in this direction by developing advanced frame-selection methods. This paper follows this direction by proposing a ``moment sampling" technique for frame selection in VideoLLMs, drawing inspiration from advancements in the moment retrieval literature. We benchmark our approach against the above-mentioned VideoQA paradigms in VideoLLMs: 1) caption-based methods that generate textual descriptions of the video followed by QA using LLMs, 2) direct frame-based methods that input sampled frames along with the text query. Our experiments demonstrate the superior effectiveness of the proposed approach.

\customparagraph{Moment Retrieval:} This video understanding task requires models to identify and localize segments of a video that are most relevant to a given text query. Early approaches tackled this alongside highlight generation~\cite{liu2022umt}. A major shift occurred with Moment-DETR~\cite{lei2021detecting}, which framed moment retrieval as a temporal object detection problem, treating relevant moments as temporal ``objects" that match the query. Inspired by the DETR~\cite{carion2020end} framework for object detection, Moment-DETR introduced a query-guided detection architecture and also proposed the QVHighlights dataset to advance research in this area.
Building on this foundation, several subsequent works have extended the DETR-based architecture for improved performance. QD-DETR~\cite{moon2023query} incorporates cross-attention between video and textual features to more effectively guide moment detection. BAM-DETR~\cite{lee2023bam} focuses on predicting boundary-oriented segments instead of center-aligned ones, addressing the inherent ambiguity in moment centers. CG-DETR~\cite{moon2023correlation} further refines the approach by modeling fine-grained correlations between question words and video clips. 
In this paper, we extend the application of moment retrieval to long-form VideoQA within the context of VideoLLMs. We leverage QD-DETR, one of the best-performing moment retrieval models, for all our experiments, integrating its outputs into a query-focused frame sampling strategy that enhances VideoLLM performance on long-form video question answering tasks.


\section{Methodology}
\label{sec:methods}

\subsection{Moment Sampling}
We propose a query-focused frame sampling strategy for VideoLLMs, termed ``Moment sampling", which leverages moment retrieval models as an alternative to the uniform sampling commonly used in prior work. Specifically, we employ the QD-DETR model~\cite{moon2023query}, a state-of-the-art moment retrieval framework trained on the QVHighlights dataset~\cite{lei2021detecting}. Given a video and its corresponding question for the VideoQA task, QD-DETR predicts a set of temporal segments, referred to as ``moments", along with associated relevance scores that indicate how relevant each moment is to the given query. These moment-level predictions form the foundation of our frame sampling strategy.

We begin by extracting moments using QD-DETR and converting their relevance scores into frame-level relevance scores. Since the predicted moments have hard boundaries, the initial relevance scores resemble a step function across the timeline. To encourage temporal smoothness, we apply Gaussian smoothing over these scores. If a frame belongs to multiple overlapping moments, its relevance score is the cumulative sum of the corresponding relevance values.
However, relying solely on raw moment predictions can introduce artifacts or lead to redundant frame selection. To overcome these limitations and enhance the informativeness and diversity of sampled frames, we introduce the following additional refinements:

\customparagraph{Quality scores:} 
    Some frames may suffer from visual degradation due to artifacts like motion blur. To down-weight such frames, we compute the variance of the Laplacian for each frame as a blur detection metric. This score is calibrated using an appropriate exponent and incorporated as a weighted penalty into the final relevance score.
 
    \customparagraph{Uniformity scores:} 
    Since some detected moments may be short or clustered closely in time, we introduce a \textit{uniformity score} to encourage temporal dispersion of sampled frames. This score is computed as the sum of squared differences between a candidate frame's timestamp and those of already selected frames, prioritizing frames that are temporally distant.
    
    \customparagraph{Frame clustering:} 
    In videos with repetitive or static scenes, visual redundancy can persist across non-adjacent frames. To mitigate this, we extract frame-level visual features and cluster them into a predefined number of groups. During selection, we ensure that only one frame is sampled from each cluster to maximize visual diversity.

\begin{figure}
    \centering
    \small
    \includegraphics[width=0.85\linewidth]{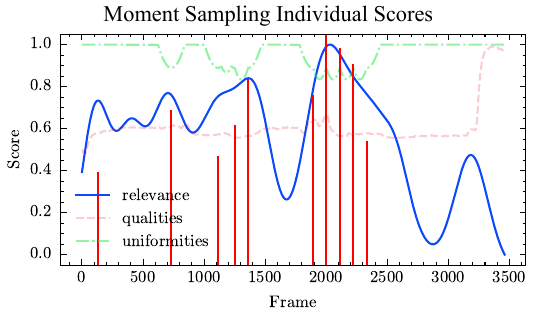}
    \caption{\textbf{
    An Example of Moment Sampling using various scores:} 
    The blue, orange, and green curves represent the relevance, quality, and uniformity scores, respectively, while the red vertical lines indicate the frames selected for sampling. The height of each red line reflects the sampling order, taller lines correspond to frames selected earlier in the process. In this example, although one moment has a significantly higher relevance score, the sampling strategy, guided by quality and uniformity scores, ensures that frames are also selected from less relevant moments. This promotes temporal diversity and robustness in the sampled set.
    }
    \label{fig:sampling}
\end{figure}

With the per-frame scores computed, we perform greedy sampling to select the final frame set. At each step, we select the frame with the highest combined score. After each frame is selected, the uniformity scores are updated to reflect its timestamp, while the relevance, quality, and clustering scores remain fixed. This process continues until the desired number of frames is sampled. The overall sampling pipeline is illustrated in Fig.~\ref{fig:sampling}.

In addition to improving the performance of VideoQA tasks, our sampling strategy provides an interpretable mechanism for temporally grounding predictions made by otherwise black-box VideoLLMs. By visualizing relevance scores and sampled frames, we offer a transparent explanation of the model's reasoning process, potentially increasing user trust and model accountability.

 \subsection{VideoLLM Setup and Evaluation}

 We experiment with a diverse set of VideoLLMs, including both proprietary models, such as GPT-4o~\cite{openai_gpt4o} and Gemini 1.5 Pro~\cite{reid2024gemini}, and open-source alternatives like VideoLLaVA~\cite{lin2023video} and InternVL2~\cite{chen2024internvl}. These models typically accept a sequence of video frames alongside a paired text prompt that contains the question to be answered.

We choose to focus on multiple-choice questions instead of open-ended ones for several compelling reasons. First, the multiple-choice format enables objective and consistent evaluation across models, minimizing the ambiguity and variability often associated with free-form responses. Second, it provides a clearer and more reliable signal of model accuracy, particularly in zero-shot settings where minor variations in phrasing can significantly affect responses. Lastly, this format simplifies downstream analysis and facilitates meaningful comparisons across different frame sampling strategies and model types.

To ensure fair and standardized evaluation, we adopt a consistent zero-shot prompting strategy. Each prompt begins with an instruction describing the task, answering a question based on the video content, followed by the question itself and a list of answer options (A to E). We explicitly instruct the model to return only the letter corresponding to the selected answer.

While most models generally comply with this output format, we observe occasional deviations where responses do not exactly match one of the provided options. In such cases, we apply a fallback mechanism: we compute the longest common subsequence (LCS) between the model’s answer and each option, selecting the one with the highest overlap.

\section{Implementation Details}
\label{sec:implementation}

All moment retrieval experiments were conducted on a single NVIDIA A10 GPU with 24GB of RAM. The open-source VideoLLMs, VideoLLaVA\cite{lin2023video} and InternVL2 (8B)~\cite{chen2024internvl}, were also run under the same GPU setup. For the proprietary models, GPT-4o from OpenAI and Gemini 1.5 Pro from Google, we only had access via their respective APIs, and thus all related inference was performed on CPU. The frame relevance sampling process was likewise executed on CPU.

We experimented with different weight combinations for the scoring components (see Fig.~\ref{fig:sampling}), and we observed strong performance with quality scores weighted with $0.5$ and uniformity scores as $2$. For quality assessment, we estimate the blur level of each frame using the variance of its Laplacian. These scores are calibrated using an exponent and combined with the other scores.
To reduce redundancy, we cluster frame-level CLIP~\cite{radford2021learning} visual features into $30$ clusters using K-means and enforce diversity by sampling at most one frame from each cluster. Relevance scores are computed as a sum of Gaussian functions centered at the predicted moment timestamps. Each Gaussian has a standard deviation equal to half the duration of the corresponding moment and is weighted by the moment’s relevance score. The final per-frame score is a weighted average of all normalized components (relevance, quality, and uniformity), and frames are selected greedily based on these scores.

\section{Experiments and Results}
\label{sec:results}

\begin{table*}
    \setlength{\tabcolsep}{5pt}
    \centering
    \small
    \caption{\textbf{Video LLMs for Long-Form VideoQA:} For each Video LLM, we report performance with and without moment sampling. For each dataset, the best performance across all models is underlined. Metric: MCQ answering accuracy (Higher is better). 
    }
    \label{tab:ours}
    \begin{tabular}{lccccccccccc}
    \toprule
    \multirow{2}{*}{Video LLM} & Moment  & EgoSchema & \multicolumn{4}{c}{CinePile} & \multicolumn{4}{c}{NextQA} & IntentQA \\
         & Sampling (Ours) & subset & AVG & CRD & NPA & TEMP & AVG & Temp & Cau & Des & full \\
         \midrule
         \multirow{2}{*}{InternVL2 (8B)} & \xmark & 51.4 & 31.6 & 32.1 & 32.9 & 26.6 & 77.7 & 75.3 & 79.4 & 76.6 & 82.4\\
         & \checkmark & \textbf{52.0} & \textbf{33.1} & \textbf{33.4} & \textbf{35.2} & \textbf{30.0} & \uline{\textbf{78.8}} & \uline{\textbf{76.9}} & \textbf{79.7} & \textbf{79.2} & \uline{\textbf{82.7}} \\
         \midrule
         \multirow{2}{*}{Gemini-1.5-pro
         } & \xmark & 66.4 & 31.7 & 38.1 & 22.8 & 28.7 & 72.8 & 67.0 & 75.9 & 74.0 & 68.1 \\
         & \checkmark & \textbf{67.8} & \textbf{33.8} & \textbf{38.4 }& \textbf{25.3} & 28.7 & \textbf{74.0} & \textbf{71.4} & 75.6 & 74.0 &  \textbf{70.0} \\
         \midrule
         \multirow{2}{*}{GPT-4o-mini} & \xmark & 56.2 & 33.1 & 37.1 & 35.4 & \textbf{27.0} & 68.6 & \textbf{64.3} & 69.8 & 74.0 & 67.7 \\
         & \checkmark & \textbf{57.6} & \textbf{34.7} & \textbf{39.1} & \uline{\textbf{36.7}} & 24.5 & \textbf{69.5} & 63.7 & \textbf{71.4} & \textbf{75.3} & \textbf{70.2} \\
         \midrule
         \multirow{2}{*}{GPT-4o} & \xmark & 72.0 & 35.2 & 40.7 & 30.4 & \uline{30.8} & \textbf{78.1} & \textbf{73.6} & \uline{\textbf{81.4}} & 75.3 & 75.8 \\
         & \checkmark & \uline{\textbf{73.6}} & \uline{\textbf{35.5}} & \uline{\textbf{41.1}} & \textbf{32.9} & \uline{30.8} & 77.4 & 72.5 & 80.4 & \textbf{76.6} & \textbf{77.3}\\
        \bottomrule
    \end{tabular}
\end{table*}

\subsection{Datasets}

We conduct experiments on four publicly available long-form VideoQA datasets, each representing a distinct domain of video content. 
More details of the datasets are as follows:

\textbf{EgoSchema}~\cite{mangalam2024egoschema} comprises egocentric videos of everyday tasks, recorded with wearable headsets as part of the Ego4D dataset. Each 3-minute video is paired with a single question, totaling 5K videos, 500 of which have publicly available answers. We conduct experiments on this answer-available subset.
Videos are accompanied by manually annotated narrations, and the question-answer pairs are generated using LLMs based on these narrations.

\textbf{CinePile}~\cite{rawal2024cinepile} consists of third-person video clips sourced from movies. The test set includes approximately 200 videos, each with an average duration of 2 minutes and 40 seconds, accompanied by around 5,000 question-answer pairs. In addition to video content, the dataset also provides subtitles and visual descriptions authored by the movie creators. The question-answer pairs are generated using large language models (LLMs), with human oversight to ensure quality.
Questions are categorized to evaluate distinct reasoning capabilities, including Character or Relationship Dynamics (CRD), Narrative or Plot Analysis (NPA), and Temporal Reasoning (TEMP). A subset of particularly challenging questions is further separated into a ``hard" split.
In our experiments, we focus exclusively on the hard split, and do not provide models with access to subtitles.

\textbf{NextQA}~\cite{xiao2021next} consists of 5,440 videos averaging 44 seconds in length, with around 52K questions split across training, validation, and test sets. As we focus on zero-shot VideoQA, we use only the validation set, which includes 570 videos. While the dataset features both open-ended and multiple-choice questions, our experiments are conducted with the multiple-choice format. The questions are categorized into three types, Temporal (Tem.), Causal (Cau.), and Descriptive (Des.), each designed to evaluate different reasoning abilities of VideoQA models.

\textbf{IntentQA}~\cite{li2023intentqa} contains around 4K videos  and 16K multiple-choice questions. These questions focus on reasoning about video intent. They use NextQA~\cite{xiao2021next} as the source dataset, identify questions based on intent and include some additional manual annotations.  
The dataset is designed so that similar actions can imply different intents depending on the context. For our experiments, we use the test set, which includes 567 videos.

\begin{table*}
    \setlength{\tabcolsep}{5pt}
    \centering
    \small
    \caption{\textbf{Comparison with captioning-based models: }
    Contrasting performance of traditional Long VideoQA pipelines (1) models that do video captioning followed by text based QA, (2) VideoLLM models that use the raw video and text question as input to produce answer, (3) frame-based Moment Sampling (MS) followed by VideoLLM based answering.
    }
    \label{tab:captions}
    \begin{tabular}{lccccccccccc}
    \toprule
    \multirow{2}{*}{Method} & LLM  & EgoSchema & \multicolumn{4}{c}{CinePile} & \multicolumn{4}{c}{NextQA} & IntentQA \\
         & Backbone & subset & AVG & CRD & NPA & TEMP & AVG & Temp & Cau & Des & full \\
         \midrule
         \multicolumn{12}{c}{Answering from Captions} \\
        \midrule
         LLoVi~\cite{zhang2024simple} &  & 57.6 & - & - & - & - & 67.7 & 61.0 & 69.5 & 75.6 & 64.0 \\
        VideoAgent~\cite{wang2024videoagent} &  & 60.2 & - & - & - & - & 71.3 & 64.5 & 72.7 & 81.1 & - \\
        \multicolumn{1}{l}{Narration + LLM } & Llama3  & 53.0 & 28.8 & 31.5 & 29.1 & 27.2 & - & - & - & - & - \\
        \multicolumn{1}{l}{VideoTree~\cite{wang2024videotree}} & GPT-4 & 66.2 & - & - & - & - & 73.5 & 67.0 & 75.2 & 81.3 & 66.9\\
        LVNet~\cite{park2024too} & GPT-4o & 66.0 & - & - & - & - & 72.9 & 65.5 & 75.0 & \textbf{81.5} & 71.1\\
        \midrule
        \multicolumn{12}{c}{Answering from Video} \\
        \midrule
        Flipped-VQA~\cite{ko2023large} &  & 44.7 & 32.5 & 36.2 & \textbf{35.6} & 23.8 & 72.0 & 69.2 & 72.7 & 75.8 & - \\
         MC-ViT-L~\cite{papalampidi2024simple} &  & 56.8 & - & - & - & - & 65.0 & - & - & - & - \\
         LongViViT~\cite{papalampidi2024simple} &  & 62.6 & - & - & - & - & - & - & - & - & - \\
        VideoLLaVA~\cite{lin2023video} & LLaVA (7B) & 20.6 & 19.3 & 18.9 & 16.5 & 23.2 & 20.5 & 17.6 & 20.6 & 27.3 & 23.3\\
        VideoLLaMA2~\cite{cheng2024videollama} & LLaMA2 & 42.2 & - & - & - & - & - & - & - & - & -\\
        Tarsier~\cite{wang2024tarsier} & Tarsier & 68.6 & - & - & - & - & - & - & - & - & 79.2 \\
        LangRepo~\cite{kahatapitiya2024language} & Mixtral (12B) & 66.2 & - & - & - & - & 60.9 & 51.4 & 64.4 & 69.1 & 59.1\\
        MoReVQA~\cite{min2024morevqa} & PALI-3 (5B) & 51.7 & - & - & - & - & 69.2 & 64.6 & 70.2 & - & -\\
        \midrule
         GPT-4o with MS (Ours)
         & GPT-4o & \textbf{73.6} & \textbf{35.5} & \textbf{41.1} & 32.9 & \textbf{30.8} & \textbf{77.4} & \textbf{72.5} & \textbf{80.4} & 76.6 & \textbf{77.3}\\       
        \bottomrule
    \end{tabular}
\end{table*}



\subsection{Quantitative Results}

Table~\ref{tab:ours} presents the quantitative comparison of various VideoLLMs with and without the proposed moment sampling strategy. Across all models and datasets, we observe a consistent performance improvement when using moment sampling over traditional uniform frame sampling. These results highlight the effectiveness of the query-focused approach in identifying semantically relevant frames. Since all evaluated datasets involve long-form videos, where crucial information is sparse and unevenly distributed, moment sampling significantly enhances performance by prioritizing frames most aligned with the question context, thereby improving both accuracy and efficiency.

%
Analyzing model performance across datasets, we can see that GPT-4o consistently outperforms others on EgoSchema and CinePile, while InternVL2 shows stronger results on NextQA and IntentQA. One possible explanation lies in the nature of the question-answer annotations: NextQA and IntentQA are manually curated by human annotators, often requiring fine-grained reasoning grounded in commonsense and real-world understanding, where InternVL2 may excel due to its training data or architecture. In contrast, EgoSchema and CinePile include question-answer pairs that are either fully or partially generated by LLMs, potentially aligning better with the style and structure of GPT-4o’s own pretraining or decoding behavior. Additionally, domain differences, such as egocentric vs. third-person perspectives and cinematic vs. daily activity, might contribute to the variation in performance across models.

Another notable observation from Table~\ref{tab:ours} is that InternVL2 consistently improves across all datasets, whereas other models exhibit occasional minor drops in performance. A likely explanation is InternVL2's greater reliance on visual content.
This indicates that models with stronger visual grounding tend to benefit more from query-focused frame selection strategies like moment sampling.


\customparagraph{Comparison with captioning-based models:}
As discussed in Sec.~\ref{sec:related}, a common alternative for handling long-form videos involves first generating textual narrations using off-the-shelf captioning or summarization models, followed by querying an LLM to answer questions based solely on this generated text. While such captioning-based methods offer scalability and reduce video processing overhead, they suffer from a critical limitation: they depend entirely on the coverage and quality of the captions. In practice, these captions often miss subtle visual cues or context-specific details that are essential for accurate question answering, particularly in complex, long-form videos.

Table~\ref{tab:captions} compare the performance of the best-performing captioning-based models with our strongest model, GPT-4o equipped with moment sampling (GPT-4o + MS) from Table~\ref{tab:ours}. The results clearly show that GPT-4o + MS consistently outperforms captioning-based approaches. This highlights the importance of preserving visual context in VideoLLMs and demonstrates that directly feeding semantically relevant frames, selected via the query-focused moment sampling strategy, enables more precise and context-aware reasoning.
Moreover, this approach offers a more faithful representation of the video’s visual semantics, allowing the model to interpret and align information more effectively with the question. Especially in long-form videos, where key information is sparse and not evenly distributed, relying on visual inputs rather than compressed textual summaries proves significantly more effective for VideoQA tasks.  

\begin{figure}[t]
    \centering
    \small
    \includegraphics[width=0.85\linewidth]{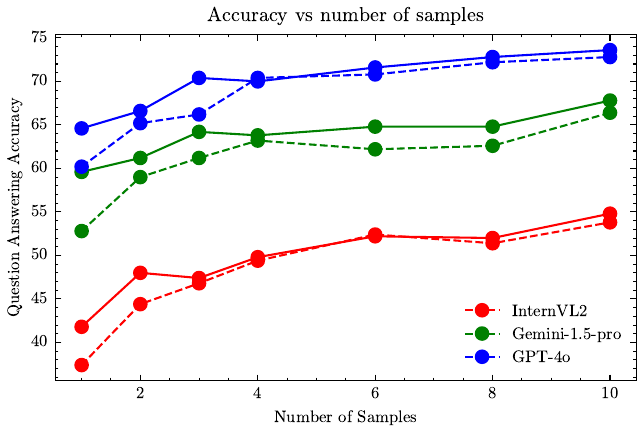}
    \caption{\textbf{Frame Sample Efficiency}. Accuracy for EgoSchema as the number of sampled frames is increased 
    with tradition uniform sampling (dashed line) and moment sampling (solid line).}
    \label{fig:scaling}
\end{figure}

\customparagraph{Frame sampling efficiency:}
Fig.~\ref{fig:scaling} analyzes the performance of selected models as the number of sampled frames is varied. Notably, the performance gap between moment sampling and uniform sampling widens as the number of frames decreases. This demonstrates that our query-based moment sampling is significantly more sample-efficient, it can often match or even surpass the performance of uniformly sampled models while using fewer frames.


%
%

\customparagraph{Visual Relience:} \label{sec:visual_reliance:} 
VideoQA tasks are fundamentally designed to evaluate a model’s ability to reason over visual content in conjunction with natural language. Thus, a high-quality VideoQA dataset should include questions that cannot be reliably answered without access to the video itself. To assess this property, we evaluate model performance with and without video input, as shown in Table~\ref{tab:visual}.

In the no-video setting, we inform the model that the question pertains to a video that is not accessible and explicitly instruct it to attempt an answer based on its prior knowledge. Across EgoSchema, NextQA, and IntentQA, we observe substantial drops in accuracy, on the order of 25-30\% points (30-50\% relative), indicating that these datasets require significant visual understanding and are less susceptible to language-only biases. In contrast, CinePile exhibits a much smaller drop (less than 6.5\% absolute, $<20\%$ relative), suggesting that many questions may be partially answerable using linguistic cues alone, perhaps due to the influence of templated or narratively suggestive phrasing.

To better isolate visual reasoning in CinePile, we further analyze settings that include or exclude subtitles (Table~\ref{tab:cinepile}). The inclusion of subtitles, which themselves often contain rich contextual information, further diminishes visual reliance. Therefore, we focus our experiments on the hard subset of CinePile without subtitles, where visual context plays a more critical role.

Moment sampling aims to enrich the quality and relevance of visual information presented to the model by prioritizing frames most aligned with the query. As such, the effectiveness of moment sampling is inherently tied to the degree to which a model depends on visual content. Models with strong visual grounding are more likely to benefit from improvements in frame selection. The performance differences observed across models in Table~\ref{tab:ours} 
can be partially attributed to the differences in visual reliance measured in Table~\ref{tab:visual}. For example, InternVL2, which shows high sensitivity to visual input, also exhibits more consistent gains with moment sampling. On the other hand, improvements are more modest for models that may rely more heavily on language priors, such as GPT-4o.

Collectively, these findings highlight the dual role of visual reliance and frame relevance in effective VideoQA. Moment sampling not only boosts sample efficiency and accuracy but also serves as a valuable strategy for enhancing multimodal alignment in long-form video understanding.

\begin{table}[t]
    \setlength{\tabcolsep}{2pt}
    \centering
    \small
    \caption{\textbf{Visual vs. Language Reliance for VideoQA.} For each dataset the performance using text alone (first row) and using vision (second row) are show). CinePile dataset perform well even without any visual information and may not be sufficiently testing the multimodal capabilities of a VideoQA model. }
    \label{tab:visual}
    \begin{tabular}{lccccc}
        \toprule
         Method & Vis & EgoSchema & CinePile & Next & Intent \\
        \midrule
        \arrayrulecolor{tablegray}
        \multirow{2}{*}{InternVL2 (8B)} & \xmark & 25.8 & 25.3 & 48.8 & 54.7 \\
         & \checkmark & 51.4 & 31.6 & 77.7 & 82.4 \\
         \midrule
        \multirow{2}{*}{Gemini-1.5-pro} & \xmark & 33.0 & 29.7 & 50.9 & 57.0 \\
         & \checkmark & 66.4 & 31.7 & 72.8 & 68.1 \\
         \midrule
        \multirow{2}{*}{GPT-4o-mini} & \xmark & 30.6 & 28.0 & 51.9 & 57.0 \\
         & \checkmark & 56.2 & 33.1 & 68.6 & 67.7 \\
         \midrule
        \multirow{2}{*}{GPT-4o} & \xmark & 42.2 & 31.1 & 52.8 & 59.3 \\
         & \checkmark & 72.0 & 37.2 & 78.1 & 75.8 \\
         \arrayrulecolor{black}
        \bottomrule
    \end{tabular}
\end{table}


\begin{table}[t]
    \setlength{\tabcolsep}{5pt}
    \centering
    \small
    \caption{\textbf{Visual Reliance on CinePile.} Visual reliance for different dataset settings.Inclusion of subtitles results in smaller loss in performance in language-only mode. However, the visual modality is more relevant when subtitles are not included.}
    \label{tab:cinepile}
    \begin{tabular}{lccccc}
        \toprule
         \multirow{2}{*}{Method} & \multirow{2}{*}{Vision} & \multicolumn{2}{c}{Subtitles} & \multicolumn{2}{c}{No Subtitles}\\
          &   & Full & Hard & Full & Hard\\
         \midrule
         \multirow{2}{*}{GPT-4o} & \xmark & 58.8 & 42.1 & 42.0 & 31.1\\
            & \checkmark & 59.7 & 43.7 & 50.4 & 37.2\\
        \bottomrule
    \end{tabular}
\end{table}


\begin{figure*}
    \centering
    \small
    \includegraphics[width=\linewidth]{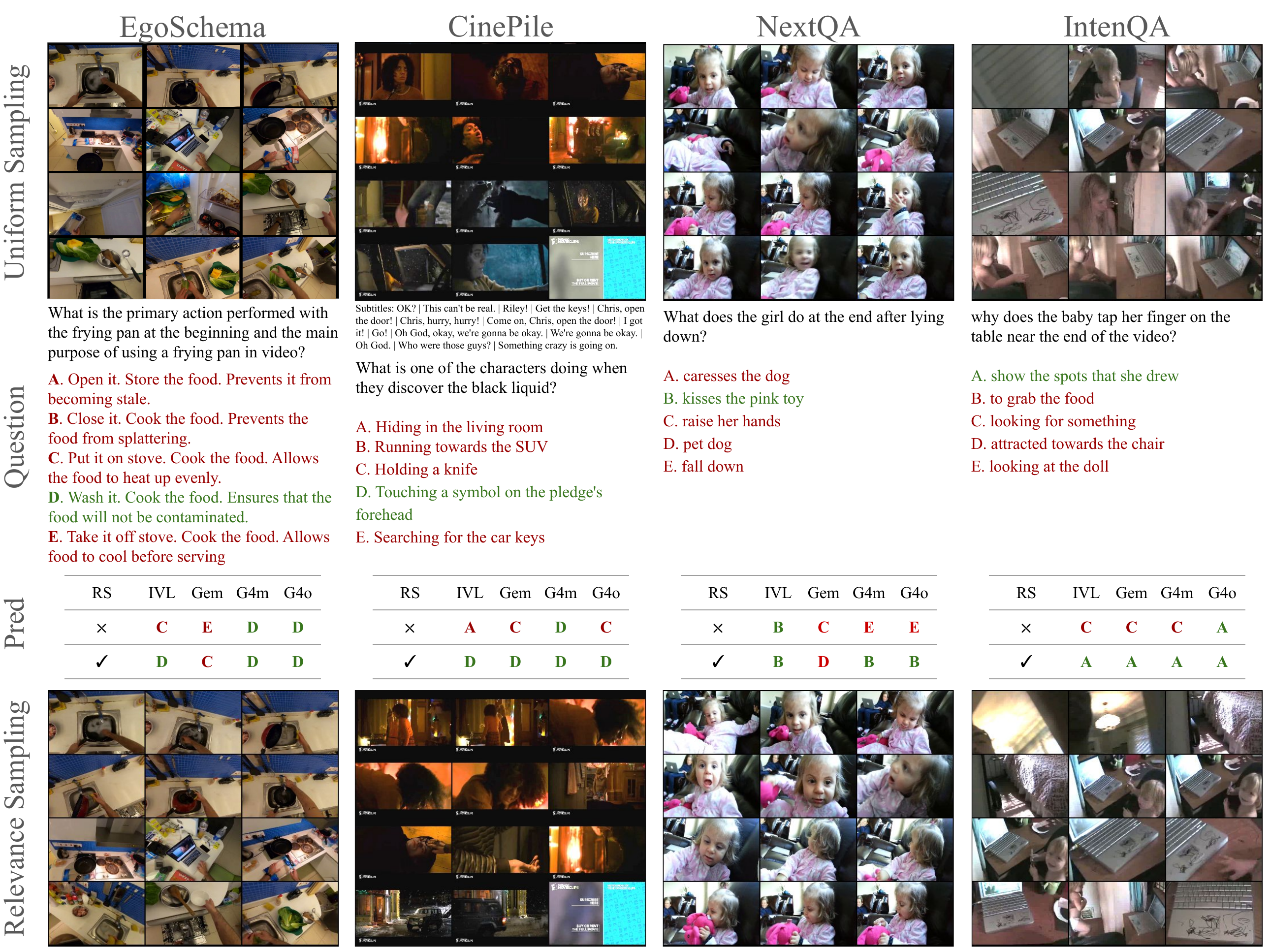}
    \caption{\textbf{Qualitative results} from the four datasets along with predictions of the VideoLLMs, with and without moment sampling. We also show collages made by uniform (top) and relevance (bottom) sampling for these questions. For each question, the correct option is in green color and incorrect options in red. Associated subtitles are shown for CinePile. }
    \label{fig:visualisations}
    \vspace{-1.5mm}
\end{figure*}

\subsection{Qualitative Results}
Fig.~\ref{fig:visualisations} shows some qualitative results showcasing model predictions along with collages of frames selected via traditional uniform sampling and our proposed moment sampling strategy. These visualizations help illustrate how moment sampling improves frame relevance and, consequently, model performance.
In the EgoSchema example, the question pertains to two key segments: the beginning of the video (washing a frying pan) and the end (using it for cooking). Moment sampling effectively allocates roughly half of the selected frames to the initial activity and the remaining half to the final event, both critical to answering the question. In contrast, uniform sampling fails to capture this temporal distribution, missing important context.
For CinePile, the answer requires detecting a specific visual cue, black liquid on a character's forehead. This detail is captured in the first frame of the penultimate row in the query-based (moment sampling) collage but only partially and less clearly in the third frame of the uniform sampling grid. The presence or absence of this frame significantly impacts the model’s ability to answer correctly.
In both NextQA and IntentQA, the questions are tied to specific incidents within the video. Uniform sampling often fails to include these moments, while moment sampling consistently captures them. For instance, in the NextQA collage, the first frame in the last row captures the key event; similarly, for IntentQA, the final frame in the penultimate row does. These contextually crucial frames are entirely missing in the corresponding uniform-sampled collages, likely leading to incorrect predictions.

These examples further highlight the core advantage of moment sampling, its ability to identify and prioritize semantically rich frames that are closely aligned with the question, particularly in long-form videos where key events may be temporally sparse and unevenly distributed.


\section{Conclusion and Future Work}
\label{sec:conclusion}
In this work, we introduced moment sampling, a relevance-driven, query-aware frame selection strategy to address the limitations of uniform sampling in long-form VideoQA in the context of VideoLLMs. By leveraging a pre-trained moment retrieval model, our method identifies video segments that are most contextually aligned with the input question, enabling VideoLLMs to focus on informative content while avoiding redundancy. Through extensive evaluation across four long-form VideoQA datasets and four state-of-the-art VideoLLMs, we demonstrated consistent performance gains. These improvements are especially notable in datasets with higher visual reliance, highlighting the importance of intelligent frame selection in multimodal reasoning tasks. Additionally, moment sampling improves sampling efficiency and offers enhanced interpretability by explicitly grounding predictions in relevant visual evidence.

Looking ahead, our work opens up several promising directions. One key avenue is end-to-end integration, where the moment retrieval and VideoQA components are trained jointly, potentially improving alignment between relevance estimation and final predictions. We also plan to explore multimodal enhancements, such as incorporating audio signals or structured video descriptions to further enrich context. Another exciting direction is temporal query decomposition, breaking complex queries into sub-questions localized to specific time segments, thereby enabling more precise and interpretable reasoning. Overall, moment sampling provides a practical foundation for scalable, efficient, and explainable long-form VideoQA in the era of VideoLLMs.

\section*{Acknowledgments and Disclosure of Funding}

The research is supported in part by grant \#2329927 from the National Science Foundation (USA).

{
    \small
    \bibliographystyle{ieeenat_fullname}
    \bibliography{CVPRW/main}

\begin{thebibliography}{61}
\providecommand{\natexlab}[1]{#1}
\providecommand{\url}[1]{\texttt{#1}}
\expandafter\ifx\csname urlstyle\endcsname\relax
  \providecommand{\doi}[1]{doi: #1}\else
  \providecommand{\doi}{doi: \begingroup \urlstyle{rm}\Url}\fi

\bibitem[Achiam et~al.(2023)Achiam, Adler, Agarwal, Ahmad, Akkaya, Aleman, Almeida, Altenschmidt, Altman, Anadkat, et~al.]{achiam2023gpt}
Josh Achiam, Steven Adler, Sandhini Agarwal, Lama Ahmad, Ilge Akkaya, Florencia~Leoni Aleman, Diogo Almeida, Janko Altenschmidt, Sam Altman, Shyamal Anadkat, et~al.
\newblock Gpt-4 technical report.
\newblock \emph{arXiv preprint arXiv:2303.08774}, 2023.

\bibitem[Apostolidis et~al.(2021)Apostolidis, Balaouras, Mezaris, and Patras]{apostolidis2021combining}
Evlampios Apostolidis, Georgios Balaouras, Vasileios Mezaris, and Ioannis Patras.
\newblock Combining global and local attention with positional encoding for video summarization.
\newblock In \emph{2021 IEEE international symposium on multimedia (ISM)}, pages 226--234. IEEE, 2021.

\bibitem[Arnab et~al.(2021)Arnab, Dehghani, Heigold, Sun, Lu{\v{c}}i{\'c}, and Schmid]{arnab2021vivit}
Anurag Arnab, Mostafa Dehghani, Georg Heigold, Chen Sun, Mario Lu{\v{c}}i{\'c}, and Cordelia Schmid.
\newblock Vivit: A video vision transformer.
\newblock In \emph{Proceedings of the IEEE/CVF international conference on computer vision}, pages 6836--6846, 2021.

\bibitem[Awadalla et~al.(2023)Awadalla, Gao, Gardner, Hessel, Hanafy, Zhu, Marathe, Bitton, Gadre, Sagawa, et~al.]{awadalla2023openflamingo}
Anas Awadalla, Irena Gao, Josh Gardner, Jack Hessel, Yusuf Hanafy, Wanrong Zhu, Kalyani Marathe, Yonatan Bitton, Samir Gadre, Shiori Sagawa, et~al.
\newblock Openflamingo: An open-source framework for training large autoregressive vision-language models.
\newblock \emph{arXiv preprint arXiv:2308.01390}, 2023.

\bibitem[Bai et~al.(2023)Bai, Bai, Yang, Wang, Tan, Wang, Lin, Zhou, and Zhou]{bai2023qwen}
Jinze Bai, Shuai Bai, Shusheng Yang, Shijie Wang, Sinan Tan, Peng Wang, Junyang Lin, Chang Zhou, and Jingren Zhou.
\newblock Qwen-vl: A frontier large vision-language model with versatile abilities.
\newblock \emph{arXiv preprint arXiv:2308.12966}, 2023.

\bibitem[Balazevic et~al.(2024)Balazevic, Shi, Papalampidi, Chaabouni, Koppula, and H{\'e}naff]{balazevic2024memory}
Ivana Balazevic, Yuge Shi, Pinelopi Papalampidi, Rahma Chaabouni, Skanda Koppula, and Olivier~J H{\'e}naff.
\newblock Memory consolidation enables long-context video understanding.
\newblock In \emph{Forty-first International Conference on Machine Learning}, 2024.

\bibitem[Carion et~al.(2020)Carion, Massa, Synnaeve, Usunier, Kirillov, and Zagoruyko]{carion2020end}
Nicolas Carion, Francisco Massa, Gabriel Synnaeve, Nicolas Usunier, Alexander Kirillov, and Sergey Zagoruyko.
\newblock End-to-end object detection with transformers.
\newblock In \emph{European conference on computer vision}, pages 213--229. Springer, 2020.

\bibitem[Chen et~al.(2023)Chen, Li, Wang, Zhao, Sun, Zhu, and Liu]{chen2023vast}
Sihan Chen, Handong Li, Qunbo Wang, Zijia Zhao, Mingzhen Sun, Xinxin Zhu, and Jing Liu.
\newblock Vast: A vision-audio-subtitle-text omni-modality foundation model and dataset.
\newblock \emph{Advances in Neural Information Processing Systems}, 36:\penalty0 72842--72866, 2023.

\bibitem[Chen et~al.(2024{\natexlab{a}})Chen, He, Li, Jin, Feng, and Liu]{chencosa}
Sihan Chen, Xingjian He, Handong Li, Xiaojie Jin, Jiashi Feng, and Jing Liu.
\newblock Cosa: Concatenated sample pretrained vision-language foundation model.
\newblock In \emph{The Twelfth International Conference on Learning Representations}, 2024{\natexlab{a}}.

\bibitem[Chen et~al.(2024{\natexlab{b}})Chen, Wu, Wang, Su, Chen, Xing, Zhong, Zhang, Zhu, Lu, et~al.]{chen2024internvl}
Zhe Chen, Jiannan Wu, Wenhai Wang, Weijie Su, Guo Chen, Sen Xing, Muyan Zhong, Qinglong Zhang, Xizhou Zhu, Lewei Lu, et~al.
\newblock Internvl: Scaling up vision foundation models and aligning for generic visual-linguistic tasks.
\newblock In \emph{Proceedings of the IEEE/CVF Conference on Computer Vision and Pattern Recognition}, pages 24185--24198, 2024{\natexlab{b}}.

\bibitem[Cheng et~al.(2024)Cheng, Leng, Zhang, Xin, Li, Chen, Zhu, Zhang, Luo, Zhao, et~al.]{cheng2024videollama}
Zesen Cheng, Sicong Leng, Hang Zhang, Yifei Xin, Xin Li, Guanzheng Chen, Yongxin Zhu, Wenqi Zhang, Ziyang Luo, Deli Zhao, et~al.
\newblock Videollama 2: Advancing spatial-temporal modeling and audio understanding in video-llms.
\newblock \emph{arXiv preprint arXiv:2406.07476}, 2024.

\bibitem[Dubey et~al.(2024)Dubey, Jauhri, Pandey, Kadian, Al-Dahle, Letman, Mathur, Schelten, Yang, Fan, et~al.]{dubey2024llama}
Abhimanyu Dubey, Abhinav Jauhri, Abhinav Pandey, Abhishek Kadian, Ahmad Al-Dahle, Aiesha Letman, Akhil Mathur, Alan Schelten, Amy Yang, Angela Fan, et~al.
\newblock The llama 3 herd of models.
\newblock \emph{arXiv preprint arXiv:2407.21783}, 2024.

\bibitem[Ging et~al.(2020)Ging, Zolfaghari, Pirsiavash, and Brox]{ging2020coot}
Simon Ging, Mohammadreza Zolfaghari, Hamed Pirsiavash, and Thomas Brox.
\newblock Coot: Cooperative hierarchical transformer for video-text representation learning.
\newblock \emph{Advances in neural information processing systems}, 33:\penalty0 22605--22618, 2020.

\bibitem[Gu and Dao(2023)]{gu2023mamba}
Albert Gu and Tri Dao.
\newblock Mamba: Linear-time sequence modeling with selective state spaces.
\newblock \emph{arXiv preprint arXiv:2312.00752}, 2023.

\bibitem[Gu et~al.(2022)Gu, Goel, and Re]{gu2022efficiently}
Albert Gu, Karan Goel, and Christopher Re.
\newblock Efficiently modeling long sequences with structured state spaces.
\newblock In \emph{International Conference on Learning Representations}, 2022.

\bibitem[Guan et~al.(2023)Guan, Pei, Shao, Liu, Li, Gu, Xu, Xu, Yan, and Lam]{guan2023pidro}
Peiyan Guan, Renjing Pei, Bin Shao, Jianzhuang Liu, Weimian Li, Jiaxi Gu, Hang Xu, Songcen Xu, Youliang Yan, and Edmund~Y Lam.
\newblock Pidro: Parallel isomeric attention with dynamic routing for text-video retrieval.
\newblock In \emph{Proceedings of the IEEE/CVF International Conference on Computer Vision}, pages 11164--11173, 2023.

\bibitem[Gygli et~al.(2014)Gygli, Grabner, Riemenschneider, and Van~Gool]{gygli2014creating}
Michael Gygli, Helmut Grabner, Hayko Riemenschneider, and Luc Van~Gool.
\newblock Creating summaries from user videos.
\newblock In \emph{Computer Vision--ECCV 2014: 13th European Conference, Zurich, Switzerland, September 6-12, 2014, Proceedings, Part VII 13}, pages 505--520. Springer, 2014.

\bibitem[Herzig et~al.(2022)Herzig, Ben-Avraham, Mangalam, Bar, Chechik, Rohrbach, Darrell, and Globerson]{herzig2022object}
Roei Herzig, Elad Ben-Avraham, Karttikeya Mangalam, Amir Bar, Gal Chechik, Anna Rohrbach, Trevor Darrell, and Amir Globerson.
\newblock Object-region video transformers.
\newblock In \emph{Proceedings of the ieee/cvf conference on computer vision and pattern recognition}, pages 3148--3159, 2022.

\bibitem[Islam et~al.(2024)Islam, Ho, Yang, Nagarajan, Torresani, and Bertasius]{islam2024video}
Md~Mohaiminul Islam, Ngan Ho, Xitong Yang, Tushar Nagarajan, Lorenzo Torresani, and Gedas Bertasius.
\newblock Video recap: Recursive captioning of hour-long videos.
\newblock In \emph{Proceedings of the IEEE/CVF Conference on Computer Vision and Pattern Recognition}, pages 18198--18208, 2024.

\bibitem[Jang et~al.(2017)Jang, Song, Yu, Kim, and Kim]{jang2017tgif}
Yunseok Jang, Yale Song, Youngjae Yu, Youngjin Kim, and Gunhee Kim.
\newblock Tgif-qa: Toward spatio-temporal reasoning in visual question answering.
\newblock In \emph{Proceedings of the IEEE conference on computer vision and pattern recognition}, pages 2758--2766, 2017.

\bibitem[Jiang et~al.(2023)Jiang, Sablayrolles, Mensch, Bamford, Chaplot, Casas, Bressand, Lengyel, Lample, Saulnier, et~al.]{jiang2023mistral}
Albert~Q Jiang, Alexandre Sablayrolles, Arthur Mensch, Chris Bamford, Devendra~Singh Chaplot, Diego de~las Casas, Florian Bressand, Gianna Lengyel, Guillaume Lample, Lucile Saulnier, et~al.
\newblock Mistral 7b.
\newblock \emph{arXiv preprint arXiv:2310.06825}, 2023.

\bibitem[Kahatapitiya et~al.(2024)Kahatapitiya, Ranasinghe, Park, and Ryoo]{kahatapitiya2024language}
Kumara Kahatapitiya, Kanchana Ranasinghe, Jongwoo Park, and Michael~S Ryoo.
\newblock Language repository for long video understanding.
\newblock In \emph{Workshop on Video-Language Models@ NeurIPS 2024}, 2024.

\bibitem[Ko et~al.(2023)Ko, Lee, Kang, Roh, and Kim]{ko2023large}
Dohwan Ko, Ji~Soo Lee, Woo-Young Kang, Byungseok Roh, and Hyunwoo~J Kim.
\newblock Large language models are temporal and causal reasoners for video question answering.
\newblock In \emph{The 2023 Conference on Empirical Methods in Natural Language Processing}, 2023.

\bibitem[Kondratyuk et~al.(2021)Kondratyuk, Yuan, Li, Zhang, Tan, Brown, and Gong]{kondratyuk2021movinets}
Dan Kondratyuk, Liangzhe Yuan, Yandong Li, Li Zhang, Mingxing Tan, Matthew Brown, and Boqing Gong.
\newblock Movinets: Mobile video networks for efficient video recognition.
\newblock In \emph{Proceedings of the IEEE/CVF conference on computer vision and pattern recognition}, pages 16020--16030, 2021.

\bibitem[Kong et~al.(2024)Kong, Goel, Badlani, Ping, Valle, and Catanzaro]{kong2024audio}
Zhifeng Kong, Arushi Goel, Rohan Badlani, Wei Ping, Rafael Valle, and Bryan Catanzaro.
\newblock Audio flamingo: A novel audio language model with few-shot learning and dialogue abilities.
\newblock In \emph{Forty-first International Conference on Machine Learning}, 2024.

\bibitem[Kuo et~al.(2023)Kuo, Piergiovanni, Kim, xiyang luo, Caine, Li, Ogale, Zhou, Dai, Chen, Cui, and Angelova]{kuo2023mammut}
Weicheng Kuo, AJ Piergiovanni, Dahun Kim, xiyang luo, Benjamin Caine, Wei Li, Abhijit Ogale, Luowei Zhou, Andrew~M. Dai, Zhifeng Chen, Claire Cui, and Anelia Angelova.
\newblock Ma{MMUT}: A simple architecture for joint learning for multimodal tasks.
\newblock \emph{Transactions on Machine Learning Research}, 2023.

\bibitem[Lee et~al.(2024)Lee, Lee, and Choi]{lee2024cast}
Dongho Lee, Jongseo Lee, and Jinwoo Choi.
\newblock Cast: cross-attention in space and time for video action recognition.
\newblock \emph{Advances in Neural Information Processing Systems}, 36, 2024.

\bibitem[Lee and Byun(2023)]{lee2023bam}
Pilhyeon Lee and Hyeran Byun.
\newblock Bam-detr: Boundary-aligned moment detection transformer for temporal sentence grounding in videos.
\newblock \emph{arXiv preprint arXiv:2312.00083}, 2023.

\bibitem[Lei et~al.(2021)Lei, Berg, and Bansal]{lei2021detecting}
Jie Lei, Tamara~L Berg, and Mohit Bansal.
\newblock Detecting moments and highlights in videos via natural language queries.
\newblock \emph{Advances in Neural Information Processing Systems}, 34:\penalty0 11846--11858, 2021.

\bibitem[Li et~al.(2023{\natexlab{a}})Li, Li, Savarese, and Hoi]{li2023blip}
Junnan Li, Dongxu Li, Silvio Savarese, and Steven Hoi.
\newblock Blip-2: Bootstrapping language-image pre-training with frozen image encoders and large language models.
\newblock In \emph{International conference on machine learning}, pages 19730--19742. PMLR, 2023{\natexlab{a}}.

\bibitem[Li et~al.(2023{\natexlab{b}})Li, Wei, Han, and Fan]{li2023intentqa}
Jiapeng Li, Ping Wei, Wenjuan Han, and Lifeng Fan.
\newblock Intentqa: Context-aware video intent reasoning.
\newblock In \emph{Proceedings of the IEEE/CVF International Conference on Computer Vision}, pages 11963--11974, 2023{\natexlab{b}}.

\bibitem[Li et~al.(2024)Li, Li, Wang, He, Wang, Wang, and Qiao]{li2024videomamba}
Kunchang Li, Xinhao Li, Yi Wang, Yinan He, Yali Wang, Limin Wang, and Yu Qiao.
\newblock Videomamba: State space model for efficient video understanding.
\newblock \emph{arXiv preprint arXiv:2403.06977}, 2024.

\bibitem[Lin et~al.(2023)Lin, Zhu, Ye, Ning, Jin, and Yuan]{lin2023video}
Bin Lin, Bin Zhu, Yang Ye, Munan Ning, Peng Jin, and Li Yuan.
\newblock Video-llava: Learning united visual representation by alignment before projection.
\newblock \emph{arXiv preprint arXiv:2311.10122}, 2023.

\bibitem[Liu et~al.(2024)Liu, Li, Wu, and Lee]{liu2024visual}
Haotian Liu, Chunyuan Li, Qingyang Wu, and Yong~Jae Lee.
\newblock Visual instruction tuning.
\newblock \emph{Advances in neural information processing systems}, 36, 2024.

\bibitem[Liu et~al.(2022)Liu, Li, Wu, Chen, Shan, and Qie]{liu2022umt}
Ye Liu, Siyuan Li, Yang Wu, Chang-Wen Chen, Ying Shan, and Xiaohu Qie.
\newblock Umt: Unified multi-modal transformers for joint video moment retrieval and highlight detection.
\newblock In \emph{Proceedings of the IEEE/CVF Conference on Computer Vision and Pattern Recognition}, pages 3042--3051, 2022.

\bibitem[Mangalam et~al.(2024)Mangalam, Akshulakov, and Malik]{mangalam2024egoschema}
Karttikeya Mangalam, Raiymbek Akshulakov, and Jitendra Malik.
\newblock Egoschema: A diagnostic benchmark for very long-form video language understanding.
\newblock \emph{Advances in Neural Information Processing Systems}, 36, 2024.

\bibitem[Min et~al.(2024)Min, Buch, Nagrani, Cho, and Schmid]{min2024morevqa}
Juhong Min, Shyamal Buch, Arsha Nagrani, Minsu Cho, and Cordelia Schmid.
\newblock Morevqa: Exploring modular reasoning models for video question answering.
\newblock In \emph{Proceedings of the IEEE/CVF Conference on Computer Vision and Pattern Recognition}, pages 13235--13245, 2024.

\bibitem[Moon et~al.(2023{\natexlab{a}})Moon, Hyun, Lee, and Heo]{moon2023correlation}
WonJun Moon, Sangeek Hyun, SuBeen Lee, and Jae-Pil Heo.
\newblock Correlation-guided query-dependency calibration in video representation learning for temporal grounding.
\newblock \emph{arXiv preprint arXiv:2311.08835}, 2023{\natexlab{a}}.

\bibitem[Moon et~al.(2023{\natexlab{b}})Moon, Hyun, Park, Park, and Heo]{moon2023query}
WonJun Moon, Sangeek Hyun, SangUk Park, Dongchan Park, and Jae-Pil Heo.
\newblock Query-dependent video representation for moment retrieval and highlight detection.
\newblock In \emph{Proceedings of the IEEE/CVF Conference on Computer Vision and Pattern Recognition}, pages 23023--23033, 2023{\natexlab{b}}.

\bibitem[OpenAI(2024)]{openai_gpt4o}
OpenAI.
\newblock Gpt-4o: A language model.
\newblock \url{https://openai.com/index/hello-gpt-4o/}, 2024.
\newblock Accessed: 2024-09-07.

\bibitem[Papalampidi et~al.(2024)Papalampidi, Koppula, Pathak, Chiu, Heyward, Patraucean, Shen, Miech, Zisserman, and Nematzdeh]{papalampidi2024simple}
Pinelopi Papalampidi, Skanda Koppula, Shreya Pathak, Justin Chiu, Joe Heyward, Viorica Patraucean, Jiajun Shen, Antoine Miech, Andrew Zisserman, and Aida Nematzdeh.
\newblock A simple recipe for contrastively pre-training video-first encoders beyond 16 frames.
\newblock In \emph{Proceedings of the IEEE/CVF Conference on Computer Vision and Pattern Recognition}, pages 14386--14397, 2024.

\bibitem[Park et~al.(2024{\natexlab{a}})Park, Kim, Ko, Kim, and Kim]{park2024videomamba}
Jinyoung Park, Hee-Seon Kim, Kangwook Ko, Minbeom Kim, and Changick Kim.
\newblock Videomamba: Spatio-temporal selective state space model.
\newblock \emph{arXiv preprint arXiv:2407.08476}, 2024{\natexlab{a}}.

\bibitem[Park et~al.(2024{\natexlab{b}})Park, Ranasinghe, Kahatapitiya, Ryu, Kim, and Ryoo]{park2024too}
Jongwoo Park, Kanchana Ranasinghe, Kumara Kahatapitiya, Wonjeong Ryu, Donghyun Kim, and Michael~S Ryoo.
\newblock Too many frames, not all useful: Efficient strategies for long-form video qa.
\newblock In \emph{Workshop on Video-Language Models@ NeurIPS 2024}, 2024{\natexlab{b}}.

\bibitem[Radford et~al.(2021)Radford, Kim, Hallacy, Ramesh, Goh, Agarwal, Sastry, Askell, Mishkin, Clark, et~al.]{radford2021learning}
Alec Radford, Jong~Wook Kim, Chris Hallacy, Aditya Ramesh, Gabriel Goh, Sandhini Agarwal, Girish Sastry, Amanda Askell, Pamela Mishkin, Jack Clark, et~al.
\newblock Learning transferable visual models from natural language supervision.
\newblock In \emph{International conference on machine learning}, pages 8748--8763. PmLR, 2021.

\bibitem[Rawal et~al.(2024)Rawal, Saifullah, Basri, Jacobs, Somepalli, and Goldstein]{rawal2024cinepile}
Ruchit Rawal, Khalid Saifullah, Ronen Basri, David Jacobs, Gowthami Somepalli, and Tom Goldstein.
\newblock Cinepile: A long video question answering dataset and benchmark.
\newblock In \emph{Synthetic Data for Computer Vision Workshop@ CVPR 2024}, 2024.

\bibitem[Reid et~al.(2024)Reid, Savinov, Teplyashin, Lepikhin, Lillicrap, Alayrac, Soricut, Lazaridou, Firat, Schrittwieser, et~al.]{reid2024gemini}
Machel Reid, Nikolay Savinov, Denis Teplyashin, Dmitry Lepikhin, Timothy Lillicrap, Jean-baptiste Alayrac, Radu Soricut, Angeliki Lazaridou, Orhan Firat, Julian Schrittwieser, et~al.
\newblock Gemini 1.5: Unlocking multimodal understanding across millions of tokens of context.
\newblock \emph{arXiv preprint arXiv:2403.05530}, 2024.

\bibitem[Team et~al.(2024)Team, Riviere, Pathak, Sessa, Hardin, Bhupatiraju, Hussenot, Mesnard, Shahriari, Ram{\'e}, et~al.]{team2024gemma}
Gemma Team, Morgane Riviere, Shreya Pathak, Pier~Giuseppe Sessa, Cassidy Hardin, Surya Bhupatiraju, L{\'e}onard Hussenot, Thomas Mesnard, Bobak Shahriari, Alexandre Ram{\'e}, et~al.
\newblock Gemma 2: Improving open language models at a practical size.
\newblock \emph{arXiv preprint arXiv:2408.00118}, 2024.

\bibitem[Touvron et~al.(2023)Touvron, Martin, Stone, Albert, Almahairi, Babaei, Bashlykov, Batra, Bhargava, Bhosale, et~al.]{touvron2023llama}
Hugo Touvron, Louis Martin, Kevin Stone, Peter Albert, Amjad Almahairi, Yasmine Babaei, Nikolay Bashlykov, Soumya Batra, Prajjwal Bhargava, Shruti Bhosale, et~al.
\newblock Llama 2: Open foundation and fine-tuned chat models.
\newblock \emph{arXiv preprint arXiv:2307.09288}, 2023.

\bibitem[Wang et~al.(2024{\natexlab{a}})Wang, Yuan, and Zhang]{wang2024tarsier}
Jiawei Wang, Liping Yuan, and Yuchen Zhang.
\newblock Tarsier: Recipes for training and evaluating large video description models.
\newblock \emph{arXiv preprint arXiv:2407.00634}, 2024{\natexlab{a}}.

\bibitem[Wang et~al.(2024{\natexlab{b}})Wang, Zhang, Zohar, and Yeung-Levy]{wang2024videoagent}
Xiaohan Wang, Yuhui Zhang, Orr Zohar, and Serena Yeung-Levy.
\newblock Videoagent: Long-form video understanding with large language model as agent.
\newblock \emph{arXiv preprint arXiv:2403.10517}, 2024{\natexlab{b}}.

\bibitem[Wang et~al.(2024{\natexlab{c}})Wang, Li, Li, Yu, He, Chen, Pei, Zheng, Xu, Wang, et~al.]{wang2024internvideo2}
Yi Wang, Kunchang Li, Xinhao Li, Jiashuo Yu, Yinan He, Guo Chen, Baoqi Pei, Rongkun Zheng, Jilan Xu, Zun Wang, et~al.
\newblock Internvideo2: Scaling video foundation models for multimodal video understanding.
\newblock \emph{CoRR}, 2024{\natexlab{c}}.

\bibitem[Wang et~al.(2024{\natexlab{d}})Wang, Yu, Stengel-Eskin, Yoon, Cheng, Bertasius, and Bansal]{wang2024videotree}
Ziyang Wang, Shoubin Yu, Elias Stengel-Eskin, Jaehong Yoon, Feng Cheng, Gedas Bertasius, and Mohit Bansal.
\newblock Videotree: Adaptive tree-based video representation for llm reasoning on long videos.
\newblock \emph{arXiv preprint arXiv:2405.19209}, 2024{\natexlab{d}}.

\bibitem[Wu et~al.(2022)Wu, Li, Mangalam, Fan, Xiong, Malik, and Feichtenhofer]{wu2022memvit}
Chao-Yuan Wu, Yanghao Li, Karttikeya Mangalam, Haoqi Fan, Bo Xiong, Jitendra Malik, and Christoph Feichtenhofer.
\newblock Memvit: Memory-augmented multiscale vision transformer for efficient long-term video recognition.
\newblock In \emph{Proceedings of the IEEE/CVF Conference on Computer Vision and Pattern Recognition}, pages 13587--13597, 2022.

\bibitem[Xiao et~al.(2021)Xiao, Shang, Yao, and Chua]{xiao2021next}
Junbin Xiao, Xindi Shang, Angela Yao, and Tat-Seng Chua.
\newblock Next-qa: Next phase of question-answering to explaining temporal actions.
\newblock In \emph{Proceedings of the IEEE/CVF conference on computer vision and pattern recognition}, pages 9777--9786, 2021.

\bibitem[Xu et~al.(2023)Xu, Ye, Yan, Shi, Ye, Xu, Li, Bi, Qian, Wang, et~al.]{xu2023mplug}
Haiyang Xu, Qinghao Ye, Ming Yan, Yaya Shi, Jiabo Ye, Yuanhong Xu, Chenliang Li, Bin Bi, Qi Qian, Wei Wang, et~al.
\newblock mplug-2: A modularized multi-modal foundation model across text, image and video.
\newblock In \emph{International Conference on Machine Learning}, pages 38728--38748. PMLR, 2023.

\bibitem[Xue et~al.(2023)Xue, Sun, Liu, Fu, Song, Li, and Luo]{xue2023clip}
Hongwei Xue, Yuchong Sun, Bei Liu, Jianlong Fu, Ruihua Song, Houqiang Li, and Jiebo Luo.
\newblock Clip-vip: Adapting pre-trained image-text model to video-language alignment.
\newblock In \emph{The Eleventh International Conference on Learning Representations}, 2023.

\bibitem[Yan et~al.(2022)Yan, Xiong, Arnab, Lu, Zhang, Sun, and Schmid]{yan2022multiview}
Shen Yan, Xuehan Xiong, Anurag Arnab, Zhichao Lu, Mi Zhang, Chen Sun, and Cordelia Schmid.
\newblock Multiview transformers for video recognition.
\newblock In \emph{Proceedings of the IEEE/CVF conference on computer vision and pattern recognition}, pages 3333--3343, 2022.

\bibitem[Yang et~al.(2023)Yang, Tian, Tan, Huang, Liu, Chang, Shi, Zhao, Bian, Wu, et~al.]{yang2023uniaudio}
Dongchao Yang, Jinchuan Tian, Xu Tan, Rongjie Huang, Songxiang Liu, Xuankai Chang, Jiatong Shi, Sheng Zhao, Jiang Bian, Xixin Wu, et~al.
\newblock Uniaudio: An audio foundation model toward universal audio generation.
\newblock \emph{arXiv preprint arXiv:2310.00704}, 2023.

\bibitem[Zhang et~al.(2024)Zhang, Lu, Islam, Wang, Yu, Bansal, and Bertasius]{zhang2024simple}
Ce Zhang, Taixi Lu, Md~Mohaiminul Islam, Ziyang Wang, Shoubin Yu, Mohit Bansal, and Gedas Bertasius.
\newblock A simple llm framework for long-range video question-answering.
\newblock In \emph{Proceedings of the 2024 Conference on Empirical Methods in Natural Language Processing}, pages 21715--21737, 2024.

\bibitem[Zhao et~al.(2023)Zhao, Misra, Kr{\"a}henb{\"u}hl, and Girdhar]{zhao2023learning}
Yue Zhao, Ishan Misra, Philipp Kr{\"a}henb{\"u}hl, and Rohit Girdhar.
\newblock Learning video representations from large language models.
\newblock In \emph{Proceedings of the IEEE/CVF Conference on Computer Vision and Pattern Recognition}, pages 6586--6597, 2023.

\bibitem[Zhu et~al.(2022)Zhu, Han, Lu, and Zhou]{zhu2022relational}
Wencheng Zhu, Yucheng Han, Jiwen Lu, and Jie Zhou.
\newblock Relational reasoning over spatial-temporal graphs for video summarization.
\newblock \emph{IEEE Transactions on Image Processing}, 31:\penalty0 3017--3031, 2022.

\end{thebibliography}
}


\end{document}